\documentclass[11pt,a4paper]{article}
\usepackage[hyperref]{emnlp-ijcnlp-2019}
\usepackage{times}
\usepackage{latexsym}

\aclfinalcopy 

\usepackage{xspace}
\usepackage{float}
\usepackage{enumerate}
\usepackage{amsmath}
\usepackage{amssymb}
\usepackage{graphicx}
\usepackage{listings}
\usepackage{color}
\usepackage{multirow}
\usepackage{array}
\usepackage{hhline}
\usepackage{nccmath}
\usepackage{diagbox}

\newcommand{\eng}[1]{\textcolor{blue}{#1}}
\newcommand{\spa}[1]{\textit{\textcolor{red}{#1}}}

\title{Language Modeling for Code-Switching:\\
	Evaluation, Integration of Monolingual Data, and Discriminative Training}

\author{Hila Gonen$^1$ \and Yoav Goldberg$^{1,2}$ \\
	$^1$Department of Computer Science,  Bar-Ilan University \\
	$^2$Allen Institute for Artificial Intelligence \\
	{\tt \{hilagnn,yoav.goldberg\}@gmail.com} \\
}

\date{}

\begin{document}
\maketitle
\begin{abstract}
  We focus on the problem of language modeling for code-switched language, in the context of automatic speech recognition (ASR). Language modeling for code-switched language is challenging for (at least) three reasons: (1) lack of available large-scale code-switched data for training; (2) lack of a replicable evaluation setup that is ASR directed yet isolates language modeling performance from the other intricacies of the ASR system; and (3) the reliance on generative modeling. We tackle these three issues: we propose an ASR-motivated evaluation setup which is decoupled from an ASR system and the choice of vocabulary, and provide an evaluation dataset for English-Spanish code-switching. This setup lends itself to a discriminative training approach, which we demonstrate to work better than generative language modeling. Finally, we explore a variety of training protocols and verify the effectiveness of training with large amounts of monolingual data followed by fine-tuning with small amounts of code-switched data, for both the generative and discriminative cases.
\end{abstract}

\section{Introduction}

This work deals with neural language modeling of code-switched language, motivated by an application to speech recognition.
Code-switching (CS) is a linguistic phenomenon defined as ``the alternation of two languages within a single discourse, sentence or constituent.'' \cite{P80}. Since CS is widely used in spoken platforms, dealing with code-switched language becomes an important challenge for automatic speech recognition (ASR) systems. To get a feeling how an ASR system trained on monolingual data performs on code-switched language, we fed the IBM English and Spanish systems\footnote{\texttt {https://www.ibm.com/watson/services/} \texttt{speech-to-text/}} with audio files of code-switched conversations from the Bangor Miami Corpus (see Section~\ref{data}). The results (examples available in Table~\ref{IBM}) exhibit two failure modes:
(1) Words and sentences in the opposite language are not recognized correctly; and (2) Code-switch points also hurt recognition of words from the main language. 
This demonstrates the need for designated speech recognition systems for CS.
A crucial component in such a CS ASR system is a strong CS language model, which is used to rank the bilingual candidates produced by the ASR system for a given acoustic signal.

\begin{table*}
	
	\begin{center}
		
		\resizebox{\textwidth}{!}{
			\begin{tabular} {  p{6.5cm} || p{6.5cm} | p{6.5cm}   }
				
				Original sentence (audio) & English model output & Spanish model output\\ \hline \hline
				
				no pero vino porque he came to check on my machine & No but we no longer he came to check on my machine & Cual provino de que en dicha c\'amara chino \\ \hline
				
				yo le invit\'e pa Easter porque me daba pena el pobre aqu\'i solo sin familia ni nada & feeling betrayed by eastern lap and I bought a new solo seem funny any now & y el envite pa\'is tampoco nada pero por aqu\'i solo sin familia ni nada\\ \hline
				
		\end{tabular}}
		\caption{\footnotesize Examples of the output of IBM's English and Spanish speech recognition systems on code-switched audio.} 
		\label{IBM}
		
	\end{center}
\end{table*}

Language models are traditionally evaluated with perplexity. However, this measure suffers from several shortcomings, in particular strong dependence on vocabulary size and lack of ability to directly evaluate scores assigned to malformed sentences. We address these deficiencies by presenting a new evaluation scheme for LM that simulates an ASR setup. Rather than requiring the LM to produce a probability for a gold sentence, we instead present the LM with a set of alternative sentences, including the gold one, and require it to rank the gold one higher. This evaluation is more realistic since it simulates the role a language model plays in the process of converting audio into text (Section~\ref{eval_method}). We create such an evaluation dataset for English-Spanish CS -- SpanglishEval (Section~\ref{eval_data}).

Additionally, LM for CS poses a unique challenge: while data for training monolingual LMs is easy to obtain in large quantities, CS occurs primarily in spoken language. This severely limits data availability, thus, small amounts of training data are an intrinsic property of CS LM. A natural approach to overcome this problem is to train monolingual language models---for which we have huge amounts of data---for each language separately, and combine them somehow into a code-switched LM. While this is relatively straightforward to do in an n-gram language model, it is not obvious how to perform such an LM combination in a non-markovian, RNN-based language model. We use a protocol for LSTM-based CS LM training which can take advantage of monolingual data \cite{CBS17} and verify its effectiveness (Section~\ref{finetune}).

Based on the new evaluation scheme we present, we further propose to learn a model for this ranking task using discriminative training. This model, as opposed to LMs, no longer depends on estimating next-word probabilities for the entire vocabulary. Instead, during training the model is introduced with positive and negative examples and is encouraged to prefer the positive examples over the negative ones. This model gives significantly better results (Section~\ref{discriminative}).

Our contributions in this work are four-fold: (a) We propose a new, vocabulary-size independent evaluation scheme for LM in general, motivated by ASR. This evaluation scheme is ranking-based and also suits CS LM; (b) We describe a process for automatically creating such datasets, and provide a concrete evaluation dataset for English-Spanish CS (SpanglishEval); (c) We present a model for this new ranking task that uses discriminative training and is decoupled of probability estimations, this model surpasses the standard LMs; (d) We verify the effectiveness of pretraining CS LM for this ranking task with monolingual data and show significant improvement over various baselines. The CS LM evaluation dataset and the code for the model are available at \texttt{https://github.com/} \texttt{gonenhila/codeswitching-lm}.

\section{Background}

\paragraph{Code-Switching}

Code-switching (CS) is defined as the use of two languages at the same discourse \cite{P80}. The mixing of different languages in various levels has been widely studied from social and linguistic point of view \cite{A99,M00,BT09}, and started getting attention also in the NLP community in the past few years \cite{SL08,AVK13,CRSC14}.

Below is an example of code-switching between Spanish and English (taken from the Bangor Miami corpus described in Section~\ref{data}). Translation to English follows:

\begin{itemize}
	\itemsep0em

	\item ``\eng{that} \spa{es su t\'io} \eng{that has lived with him like I don't know how like} \spa{ya} \eng{several years}..." \\
	that his uncle who has lived with him like, I don't know how, like several years already...
	
\end{itemize}

Code-switching is becoming increasingly popular, mainly among bilingual communities. Yet, one of the main challenges when dealing with CS is the limited data and its unique nature: it is usually found in non standard platforms such as spoken data and social media and accessing it is not trivial \cite{CST16}.

\paragraph{Shortcomings of Perplexity-based Evaluation for LM}

The most common evaluation measure of language models is perplexity. Given a language model $M$ and a test sequence of words $w_1, ..., w_N$, the perplexity of $M$ over the sequence is defined as:

\[ \scalebox{1.2} {    $   2^  {   -\frac{1}{N} \sum_{i=1}^{N} {\log_2 M(w_i)}   }    $    } \]
where $M(w_i)$ is the probability the model assigns to $w_i$.

A better model is expected to give higher probability to sentences in the test set, that is, lower perplexity. However, this measure is not always well aligned with the quality of a language model as it should be. For example, Tran et al. \shortcite{TBM18} show that RNNs capture long-distance syntactic dependencies better than attention-only LMs, despite having higher (worse) perplexity. Similarly, better perplexities often do not translate to better word-error-rate (WER) scores in an ASR system \cite{HL18}.

This highlights a shortcoming of perplexity-based evaluation: the method is rewarded for assigning high probability to gold sentences, but is not directly penalized for assigning high probability to highly implausible sentences. When used in a speech recognition setup, the LM is expected to do just that: score correct sentences above incorrect hypotheses.

Another shortcoming of perplexity-based evaluation is that it requires the compared models to have the same support (in other words, the same vocabulary). Simply adding words to the vocabulary, even if no additional change is done to the model, will likely result in higher perplexity for the same dataset. It is also sketchy to compare perplexities of word-based LMs and character-based LMs for the same reason.

\paragraph{Problem with WER-based evaluation}
Evaluating LM models based on the final WER of an ASR system side-steps these two issues: it evaluates the LM on incorrect sentences, and seamlessly compares LMs with different support. However, this makes the evaluation procedure highly dependent on a particular ASR system. This is highly undesirable, as ASR systems are hard to set up and tune and are not standardized. This both conflates the LM performance with performance of other aspects of the ASR system, and makes it hard to replicate the evaluation procedure and fairly compare results across different publications. Indeed, as discussed in Section~\ref{related}, most previous works on CS LM use an ASR system as part of their evaluation setup, and none of them compares to any other work. Moreover, no standard benchmark or evaluation setup exists for CS LM.

\section{An Alternative Evaluation Method}
\label{eval_method}

We propose an evaluation technique for language models which is ASR-system-independent, that does take into account incorrect sentences and allows to compare LMs with different support or OOV handling techniques.

We seek a method that meets the following requirements: (1) Prefers language models that prioritize correct sentences over incorrect ones; (2) Does not depend on the support (vocabulary) of the language model; (3) Is independent of and not coupled with a speech recognition system.

To meet these criteria, we propose to assemble a dataset comprising of gold sentences, where each gold sentence is associated with a set of alternative sentences, and require the LM to identify the gold sentence in each set. The alternative sentences in a set should be related to the gold sentence. Following the ASR motivation, we take the alternative set to contain sentences that are phonetically related (similar-sounding) to the gold sentence.
This setup simulates the task an LM is intended to perform as a part of an ASR system: given several candidates, all originating from the same acoustic signal, the LM should choose the correct sentence over the others. 

\paragraph{A New Evaluation metric} 
Given this setup, we propose to use the \emph{accuracy} metric: the percentage of sets in which the LM (or other method) successfully identified the gold sentence among the alternatives.\footnote{We chose accuracy over WER as the default metric since in our case, WER should be treated with caution: the alternatives created might be ``too close" to the gold sentence (e.g. when only a small fraction of the gold sentence is sampled and replaced) or ``too far" (e.g. a Spanish alternative for an English sentence), thus affecting the WER.} The natural way of using an LM for identifying the gold sentences is assigning a probability to each sentence in the set, and choosing the one with highest probability. Yet, the scoring mechanism is independent of perplexity, and addresses the two deficiencies of perplexity based evaluation discussed above.

Our proposed evaluation method is similar in concept to the NMT evaluation proposed by Sennrich \shortcite{S16}. There, the core idea is to measure whether a reference translation is more probable under an NMT model than a contrastive translation which introduces a specific type of error.

\section{Evaluation Dataset}
\label{eval_data}

We now turn to construct such an evaluation dataset.
One method of obtaining sentence-sets is feeding acoustic waves into an ASR system and tracking the resulting lattices. However, this requires access to the audio signals, as well as a trained system for the relevant languages. We propose instead a generation method that does not require access to an ASR system.

The dataset we create is designed for evaluating English-Spanish CS LMs, but the creation process can be applied to other languages and language pairs.\footnote{The requirements for creating an evaluation dataset for a language pair L1,L2 is to have access to code-switched sentences where each word is tagged with a language ID (language ID is not mandatory, but helps in cases in which a word is found in the vocabulary of both languages and is pronounced differently), compatible pronunciation dictionaries and unigram word probabilities for each of the languages.}

The process of alternative sentence creation is as follows: (1) Convert a sequence of language-tagged words (either CS or monolingual) into a sequence of the matching phonemes (using pronunciation dictionaries); (2) Decode the sequence of phonemes into new sentences, which include words from either language (possibly both); (3) When decoding, allow minor changes in the sequence of phonemes to facilitate the differences between the languages.

These steps can be easily implemented using composition of finite-state transducers (FSTs).\footnote{Specifically, we compose the following FSTs: (a)	an FST for converting a sentence into a sequence of phonemes, (b) an FST that allows minor changes in the phoneme sequence, (c) an FST for decoding a sequence of phonemes into a sentence, the inverse of (a).}	

For each gold sentence (which can be either code-switched or monolingual) we create alternatives of all three types: (a) code-switched sentences, (b) sentences in L1 only, (c) sentences in L2 only.

We created such a dataset for English-Spanish with gold sentences from the Bangor Miami Corpus (Section~\ref{data}). Figure~\ref{ex} shows two examples from the dataset. In each, the gold sentence is followed by a single code-switched alternative, a single English alternative, and a single Spanish one (a subset of the full set).

	\begin{figure}
		
		\fbox{\parbox{7.5cm}{			
				
				Example 1:
				
				\hspace{0.3cm} Gold: \spa{pero} \eng{i thought} \spa{ella fue} \eng{like very good} 
				
				\hspace{0.3cm} Alt-CS: \spa{pero} \eng{i thought a} \spa{llev\'o} \eng{alike very good} 
				
				\hspace{0.3cm} Alt-EN: \eng{peru i thought a of alike very could} 
				
				\hspace{0.3cm} Alt-SP: \spa{pero azote la fue la que rico}

				Example 2:
				
				\hspace{0.3cm} Gold: \spa{vamos a ser juntos} \eng{twenty four seven} 
				
				\hspace{0.3cm} Alt-CS: \spa{vamos a ser juntos} \eng{when de for} \spa{saben} 
				
				\hspace{0.3cm} Alt-EN: \eng{follows a santos twenty for seven} 
				
				\hspace{0.3cm} Alt-SP: \spa{vamos hacer junto sent\'i for saben} 
				
		}}

		\caption{Examples from SpanglishEval -- the English-Spanish evaluation dataset. The first sentence in each example is the gold sentence, followed by a generated code-switched alternative, a generated English alternative, and a generated Spanish one.  Blue (normal) marks English, while red (italic) marks Spanish.}
		
		\label{ex}
		
	\end{figure}

\subsection{Technical details}

When creating code-switched alternatives, we want to encourage the creation of sentences that include both languages, and that differ from each other. This is done with scores determined by some heuristics, such as preferring sentences that include more words from the language that was less dominant in the original one and vice versa. As part of this we also try to avoid noise from addition of short frequent words, by encouraging the average word length to be high\footnote{The implementation of these heuristics is part of our code that is available online.}. We create 1000-best alternatives from the FST, re-score them according to the heuristic and keep the top 10.

We discard sets in which the gold sentence has less than three words (excluding punctuation), and also sets with less than 5 alternatives in one of the three types.

We randomly choose 250 sets in which the gold sentence is code-switched, and 750 sets in which the gold sentence is monolingual, both for the development set and for the test set. This percentage of CS sentences is higher than in the underlying corpus in order to aid a meaningful comparison between the models regarding their ability to prefer gold CS sentences over alternatives. The statistics of the dataset are detailed in Table~\ref{stats_data}.

Further details regarding the implementation can be found in the Appendix.

\begin{table}
	
	\begin{center}
		\resizebox{0.48\textwidth}{!}{
			\begin{tabular} {  l || l| l   }
				
				& dev set & test set \\ \hline \hline
				
				no. of sets & 1000 & 1000 \\\hline
				no. of sentences & 30884 & 30811 \\\hline
				no. of CS alternatives & 10000 & 10000 \\\hline
				no. of English alternatives & 9999  &  9993 \\\hline
				no. of Spanish alternatives & 9885  & 9818 \\\hline
				no. of CS gold sentences & 250/1000 & 250/1000 \\\hline
				
			\end{tabular}
		}
		
		\caption{\footnotesize Statistics of the dataset.} 
		
		\label{stats_data}
		
	\end{center}
\end{table}

\section{Using Monolingual Data for CS LM}
\label{finetune}

Data for code-switching is relatively scarce, while monolingual data is easy to obtain. The question then is \emph{how do we efficiently incorporate monolingual data when training a CS LM?}

We use an effective training protocol (\textsc{FineTuned}) for incorporating monolingual data into the language model, similar to the best protocol introduced in \cite{CBS17}. We first pre-train a model with monolingual sentences from both English and Spanish (with shared vocabulary for both languages). This essentially trains two monolingual models, one for English and one for Spanish, but with full sharing of parameters. Note that in this pre-training phase, the model is not exposed to any code-switched sentence. 

We then use the little amount of available code-switched data to further train the model, making it familiar with code-switching examples that mix the two languages. This fine-tunning procedure enables the model to learn to correctly combine the two languages in a single sentence. 

We show in Section~\ref{results} that adding the CS data only at the end, in the described manner, works substantially better than several alternatives, verifying the results from \cite{CBS17}.

\section{Discriminative Training}
\label{discriminative}

Our new evaluation method gives rise to training models that are designated to the ranking task. As our main goal is to choose a single best sentence out of a set of candidate sentences, we can focus on training models that score whole sentences with discriminative training, rather than using the standard probability setting of LMs. Discriminative approach unbinds us from the burden of estimating probability distributions over all the words in the vocabulary and allows us to simply create representations of sentences and match them with scores.

Using negative examples is not straight forward in LM training, but is essential for speech recognition, where the language model needs to distinguish between ``good" and ``bad" sentences. Our discriminative training is based on the idea of using both positive and negative examples during training. The training data we use is similar in nature to that of our test data: sets of sentences in which only a single one is a genuine example collected from a real CS corpus, while all the others are synthetically created. 

During training, we aim to assign the gold sentences with a higher score than the scores of the others. For each sentence, we require the difference between its score and the score of the gold sentence to be as large as its WER with respect to the gold sentence. This way, the farther a sentence is from the gold one, the lower its score is.  

Formally, let $s_1$ be the gold sentence, and $s_2, ..., s_m$ be the other sentences in that set. The loss of the set is the sum of losses over all sentences, except for the gold one: $$ \sum_{i=2}^m {\max(0, \textrm{WER}(s_1, s_i) - [\textrm{score}(s_1) - \textrm{score}(s_i)])}$$
where $\textrm{score}(s_i)$ is computed by the multiplication of the representation of $s_i$ and a learned vector $w$: $$ \textrm{score}(s_i) = w \cdot \textrm{repr}(s_i)$$

A sentence is represented with its BiLSTM representation -- the concatenation of the final states of forward and backward LSTMs. Formally, a sentence $s = w_1, ..., w_n$ is represented as follows:
$$ \textrm{repr}(s) = LSTM(w_1, ..., w_n) \circ LSTM(w_n, ..., w_1)$$
where $\circ$ is the concatenation operation.	

\paragraph{Incorporating monolingual data}
In order to use monolingual data in the case of discriminative training, we follow the same protocol: as a first step, we create alternative sentences for each monolingual sentence from the monolingual corpora. We train a model using this data, and as a next step, we fine-tune this model with the sets of sentences that are created from the CS corpus.

\section{Empirical Experiments}

\paragraph{Models and Baselines}

We report results on two models that use  discriminative training: \textsc{CS-only-discriminative} only trains on data that is created based on the small code-switched corpus, while \textsc{Fine-Tuned-discriminative} first trains on data created based on the monolingual corpora and is then fine-tuned using the data created from the code-switched corpus.

We compare our models to several baselines, all of which use standard LM training: \textsc{EnglishOnly-LM} and \textsc{SpanishOnly-LM} train on the monolingual data only. Two additional models train on a combination of the code-switched corpus and the two monolingual corpora: the first (\textsc{All:Shuffled-LM}) trains on all sentences (monolingual and code-switched) presented in a random order. The second (\textsc{All:CS-last-LM}) trains each epoch on the monolingual datasets followed by a pass on the small code-switched corpus. The models \textsc{CS-only-LM} and  \textsc{Fine-Tuned-LM} are the equivalents of \textsc{CS-only-discriminative} and  \textsc{Fine-Tuned-discriminative} but with standard LM training. 

\paragraph{Code-switching corpus}
\label{data}

We use the Bangor Miami Corpus, consisting of transcripts of conversations by Spanish-speakers in Florida, all of whom are bilingual in English.\footnote{\texttt {http://bangortalk.org.uk/speakers.} \texttt{php?c=miami}}

We split the sentences (45,621 in total) to train/dev/test with ratio of 60/20/20 respectively, and evaluate perplexity on the dev and test sets. 

The dataset described in Section~\ref{eval_data} is based on sentences from the dev and test sets, and serves as our main evaluation method.

\paragraph{Monolingual corpora}

The monolingual corpora used for training the English and Spanish monolingual models are taken from the OpenSubtitles2018 corpus~\cite{T09},\footnote{\texttt{http://opus.nlpl.eu/} \texttt {OpenSubtitles2018.php}\\\texttt{http://www.opensubtitles.org/}} of subtitles of movies and TV series.

We use 1M lines from each language, with a split of 60/20/20 for train/dev/test, respectively. The test set is reserved for future use. For discriminative training we use 1/6 of the monolingual training data (as creating the data results in roughly 30 sentences per gold sentence).

Additional details on preprocessing and statistics of the data can be found in the Appendix.

\paragraph{Training}

We implement our language models in DyNet \cite{NDG17}. Our basic configuration is similar to that of Gal and Ghahramani \shortcite{GAG16} with minor changes. It has a standard architecture of a 2-layer LSTM followed by a softmax layer, and the optimization is done with SGD (see Appendix for details).

Tuning of hyper-parameters was done on the PTB corpus, in order to be on par with state-of-the-art models such as that of Merity et al. \shortcite{MKS17}. We then trained the same LM on our CS corpus with no additional tuning and got perplexity of 44.06, better than the 52.99 of Merity et al. \shortcite{MKS17} when using their default parameters on the CS corpus.\footnote{\texttt{https://github.com/salesforce/} \texttt{awd-lstm-lm}} We thus make no further tuning of hyper-parameters. 

When changing to discriminative training, we perform minimal necessary changes: discarding weight decay and reducing the learning rate to 1. The weight vector in the discriminative setting is learned jointly with the other parameters of the network.

As done in previous work, in order to be able to give a reliable probability to every next-token in the test set, we include all the tokens from the test set in our vocabulary and we do not use the ``$<$unk$>$" token. We only train those that appear in the train set. Handling of unknown tokens becomes easier with discriminative training, as scores are assigned to full sentences. However, we leave this issue for future work.

\section{Results}
\label{results}

The results of the different models are presented in Table~\ref{res1}. For each model we report both perplexity and accuracy (except for discriminative training, where perplexity is not valid), where each of them is reported according to the best performing model on that measure (on the dev set). We also report the WER of all models, which correlates perfectly with the accuracy measure.

\begin{table}[t!]
	
	\begin{center}
		
		\resizebox{0.48\textwidth}{!}{
			
			\begin{tabular} {  l | c | c | c ||  c|c|c }

				& \multicolumn{3}{c||}{dev} & \multicolumn{3}{c}{test} \\  \hline

				& perp $\downarrow$ & acc $\uparrow$ & wer $\downarrow$ & perp $\downarrow$ & acc $\uparrow$ & wer $\downarrow$ \\\hline \hline

				\textsc{Spanish-only-LM} 
				& 329.68 & 26.6 & 30.47 & 322.26 & 25.1 & 29.62 \\\hline						
				
				\textsc{English-only-LM} 
				& 320.92 & 29.3 & 32.02 & 314.04 & 30.3 & 32.51 \\\hline
				
				\textsc{All:CS-last-LM} 
				& 76.64 & 47.8 & 14.56 & 76.97 & 49.2 & 14.13 \\\hline
				
				\textsc{All:Shuffled-LM} 
				& 68.00 & 51.8 & 13.64 & 68.72 & 51.4 & 13.89 \\\hline
				
				\textsc{CS-only-LM} 
				& 43.20 & 60.7 & 12.60 & 43.42 & 57.9 & 12.18  \\\hline
				
				\textsc{CS-only+vocab-LM} 
				& 45.61 & 61.0 & 12.56 &  45.79 & 58.8 & 12.49 \\\hline
				
				\textsc{Fine-Tuned-LM} 
				& 39.76 & 66.9 & 10.71 &  40.11 & 65.4 & 10.17 \\\hline \hline
				
				\textsc{CS-only-disc} 
				& -- & 72.0 & 6.35 &  -- & 70.5  & 6.70 \\\hline
				
				\textsc{Fine-Tuned-disc} 
				& -- & \textbf{74.2} & \textbf{5.85} & -- & \textbf{75.5} & \textbf{5.59} \\\hline
				
			\end{tabular}
		}
		
		\caption{\footnotesize Results on the dev set and on the test set. ``perp" stands for perplexity, ``acc" stands for accuracy (in percents), and ``wer" stands for word-error-rate.} 
		
		\label{res1}
	
	\end{center}
\end{table}

\subsection{Using monolingual data}

As mentioned above, both in standard LM and in discriminative training, using monolingual data in a correct manner (\textsc{Fine-Tuned-LM} and \textsc{Fine-Tuned-discriminative}) significantly improves over using solely the code-switching data. In standard LM, adding monolingual data results in an improvement of 7.5 points (improving from an accuracy of 57.9\% to 65.4\%), and in the discriminative training it results in an improvement of 5 points (improving from an accuracy of 70.5\% to 75.5\%). Even though both \textsc{All:shuffled-LM} and \textsc{All:CS-last-LM} use the same data as the \textsc{Fine-Tuned-LM} model, they perform even worse than \textsc{CS-only-LM} that does not use the monolingual data at all. This emphasizes that the manner of integration of the monolingual data has a very strong influence. 

Note that the \textsc{Fine-Tuned-LM} model also improves perplexity. As perplexity is significantly affected by the size of the vocabulary---and to ensure fair comparison---we also add the additional vocabulary from the monolingual data to \textsc{CS-only-LM} (\textsc{CS-only+vocab-LM}). 
Extending the vocabulary without training those additional words, results in a 2.37-points loss on the perplexity measure, while our evaluation metric (accuracy) stays essentially the same. This demonstrates the utility of our proposed evaluation compared to using perplexity, allowing it to fairly compare models with different vocabulary sizes.

In order to examine the contribution of the monolingual data, we also experimented with subsets of the code-switching training data. Table~\ref{res2} depicts the results when using subsets of the CS training data with discriminative training. The less code-switching data we use, the more the effect of using the monolingual data is significant: we gain 8.8, 6.5, 4.2 and 5 more accuracy points with 25\%, 50\%, 75\% and 100\% of the data, respectively. In the case of 25\% of the data, the \textsc{Fine-Tuned-discriminative} model improves over \textsc{CS-only-discriminative} by 17 relative percents. 

\subsection{Standard LMs vs. Discriminative Training}

In the standard training setting, The \textsc{Fine-Tuned-LM} baseline is the strongest baseline, outperforming all others with an accuracy of 65.4\%. Similarly, when using discriminative training, the \textsc{Fine-Tuned-discriminative} model outperforms the \textsc{CS-only-discriminative} model. Note that using discriminative training, even with no additional monolingual data, leads to better performance than that of the best language model: the \textsc{CS-only-discriminative} model achieves an accuracy of 70.5\%, 5.1 points more than the accuracy of the \textsc{Fine-Tuned-LM} model. We gain further improvement by adding monolingual data and get an even higher accuracy of 75.5\%, which is 10.1 points higher than the best language model.

\begin{table}
	\begin{center}

		\resizebox{0.48\textwidth}{!}{
		\begin{tabular} { l ||  
				c|c ||
				c|c ||
				c|c ||
				c|c  }	
			
			& \multicolumn{2}{c||}{25\% train}
			& \multicolumn{2}{c||}{50\% train}
			& \multicolumn{2}{c||}{75\% train}
			& \multicolumn{2}{c}{full train} 
			\\ \hline 
			
			& dev & test & dev & test & dev & test & dev & test \\ \hline \hline
			
			\textsc{CS-only}
			& 58.4 & 58.9 & 65.2 & 63.6 & 70.8 & 68.8 & 72.0 & 70.5 \\\hline
			
			\textsc{Fine-Tuned}
			& \textbf{68.4} & \textbf{67.7} & \textbf{71.9} & \textbf{70.1} & \textbf{72.8} & \textbf{73.0} & \textbf{74.2} & \textbf{75.5} \\\hline
			
		\end{tabular}}

		\caption{\footnotesize Results on the dev set and on the test set using discriminative training with only subsets of the code-switched data.} 
		
		\label{res2}
		
	\end{center}
\end{table}

\paragraph{Limitations}
	Our evaluation setup in the discriminative training case is not ideal: the negative samples in both the train and test sets are artificially created by the same mechanism. Thus, high performance on the test set in the discriminative case may result from ``leakage'' in which the model learns to rely on idiosyncrasies and artifcats of the data generation mechanism. As such, these results may not transfer as is to a real-world ASR scenario, and should be considered as an optimistic estimate of the actual gains.\footnote{An ideal evaluation setup will use an artificial dataset for training, and a dataset obtained from an acoustic model of a code-switched ASR system for testing. However, such an evaluation setup requires access to a high-quality code-switched acoustic model, which we do not posses nor have the means to obtain. Furthermore, it would also tie the evaluation to a specific ASR system, which we aim to avoid.} Nevertheless, we do believe that the accuracy improvements of the discriminative setup are real, and should translate to improvements also in the real-world scenario.
	
	To alleviate the concerns to some extent, we consider the case in which the leakage is in the induced negative words distribution.\footnote{As our generation process, much like an acoustic model, makes only local word replacements, the choice of replacement words is the most salient difference from a real acoustic model, and has the largest leakage potential.} We re-train the discriminative model using a BOW representation of the sentence. This results in accuracies of 52.1\% and 47.4\% for dev and test, respectively, far below those of the basic model. A model that leaks in word distribution would score much higher in this evaluation, indicating that the improvement is likely not due to this form of leakage, and that most of the improvement of the discriminative training is likely to translate to a real ASR scenario.

\section{Analysis}

Table~\ref{250} breaks down the results of the different models according to two conditions: when the gold sentence is code-switched, and when the gold sentence is monolingual. 

As expected, the \textsc{Fine-Tuned-discriminative} model is able to prioritize the gold sentence better than all other models, under both conditions. The improvement we get is most significant when the gold sentence is CS: in those cases we get a dramatic improvement of 27.73 accuracy points (a relative improvement of 58\%). Note that for the standard LMs, the cases in which the gold sentence is CS are much harder, and they perform badly on this portion of the test set. However, using discriminative learning enable us to get improved performance on both portions of the test set and to get comparable results on both parts.  

\begin{table}
	
	\begin{center}
		\resizebox{0.48\textwidth}{!}{
			
			\begin{tabular} { l ||c|c || c|c  }
				
				& \multicolumn{2}{c||}{dev} & \multicolumn{2}{c}{test} \\ \hline 
				& CS & mono & CS & mono  \\ \hline

				\textsc{CS-only-LM} 
				& 45.20 & 65.87 & 43.20 & 62.80 \\\hline
				
				\textsc{Fine-Tuned-LM} 
				& 49.60 & 72.67 & 47.60 & 71.33 \\\hline \hline
				
				\textsc{CS-only-disc} 
				& \textbf{75.60} & 70.40 & 70.80 & 70.53 \\\hline					
				
				\textsc{Fine-Tuned-disc} 
				& 70.80 & \textbf{74.40} & \textbf{75.33} & \textbf{75.87} \\\hline

			\end{tabular}
		}
		\caption{\footnotesize Accuracy on the dev set and on the test set, according to the type of the gold sentence in the set: code-switched (CS) vs. monolingual (mono).} 
		\label{250}
		
	\end{center}
\end{table}

A closer examination of the mistakes of the \textsc{Fine-Tuned-LM} and \textsc{Fine-Tuned-discriminative} models reveals the superiority of the discriminative training in various cases. Table~\ref{mistakes} presents several examples in which \textsc{Fine-Tuned-LM} prefers a wrong sentence whereas \textsc{Fine-Tuned-discriminative} identifies the gold one. In examples 1--4 the gold sentence was code-switched but \textsc{Fine-Tuned-LM} forced an improbable monolingual one. Examples 5 and 6 show mistakes in monolingual sentences.

While discriminative training is significantly better than the standard LM training, it can still be improved quite a bit. Table~\ref{mistakes2} lists some of the mistakes of the \textsc{Fine-Tuned-discriminative} model: in examples 1, 2 and 3, the gold sentence was code-switched but the model preferred a monolingual one, in example 4 the model prefers a wrong CS sentence over the gold monolingual one, and in 5 and 6 the model makes mistakes in monolingual sentences.

\begin{table*}
	\begin{center}
		\resizebox{\textwidth}{!}{
			\begin{tabular} {  l | l | l | l  }
				\hline
				no. & type & Gold sentence & Choice of \textsc{Fine-Tuned-LM} model \\ \hline \hline
				
				1 & CS $\rightarrow$ mono & \eng{and in front of everybody} \spa{me salt\'o} . & \eng{and in front of everybody muscle too} . \\ \hline	
				
				2 & CS $\rightarrow$ mono & \spa{porque} \eng{the sewer system has them in there} \spa{porque} . & \spa{porque de ser sistemas de mil de porque} . \\ \hline			
				
				3 & CS $\rightarrow$ mono & \spa{entonces} \eng{what i had done is gone ahead and printed it out} . & \eng{and all says what i had done is gone ahead and printed it out} .		
				\\ \hline
				
				4 & CS $\rightarrow$ mono & \spa{es que creo que quedan como novecientos oportunidades para} \eng{beta testers} . & \spa{es que creo que que del como novecientos oportunidad esperaba de estar} .
				\\ \hline	
				
				5 & mono $\rightarrow$ mono & \eng{we we stop beyond getting too cruel} . & 
				\eng{we we stop be and getting to cruel} .
				\\ \hline
				
				6 & mono $\rightarrow$ mono & \spa{en mi casa tengo tanto huevo duro} . & \spa{en mi casa tengo tanto a futuro} .
				\\ \hline

			\end{tabular}
		}
		\caption{\footnotesize Examples of sentences the \textsc{Fine-Tuned-discriminative} model identifies correctly while the \textsc{Fine-Tuned-LM} model does not.}
		\label{mistakes}
		
	\end{center}
\end{table*}

\begin{table*}
	
	\begin{center}
		\resizebox{\textwidth}{!}{
			\begin{tabular} { l | l |l | l }
				\hline
				no. & type & Gold sentence & Choice of \textsc{Fine-Tuned-discriminative} model \\ \hline \hline

				1 & CS $\rightarrow$ mono & \spa{que} \eng{by the way} \spa{se vino ilegal} . & 		
				\spa{que va de esa vino ilegal} . \\ \hline
				
				2 & CS $\rightarrow$ mono & \spa{son un} \eng{website there} .&
				\eng{so noon website there} .\\ \hline
				
				3 & CS $\rightarrow$ mono & \eng{no i have never felt nothing close to the} \spa{esp\'iritu santo} \eng{never} .&
				\eng{no i have never felt nothing close to the a spirits on to never} .\\ \hline
				
				4 & mono $\rightarrow$ CS & \spa{T\'u sabes que que el cuerpo empieza a sentirse raro} . & \spa{T\'u sabes que que el cuerpo empieza sentir ser} \eng{arrow} .
				\\ \hline
				
				5 & mono $\rightarrow$ mono & \spa{bueno ya son las las y cuarenta casi casi} . & \spa{bueno ya son las lassie cuarenta que si casi} .
				\\ \hline
				
				6 & mono $\rightarrow$ mono & \eng{we have a peter lang} . &
				\eng{we have a bitter lung} .\\ \hline

			\end{tabular}
		}
		\caption{\footnotesize Examples of sentences that the \textsc{Fine-Tuned-discriminative} model fails to identify.} 
		\label{mistakes2}
		
	\end{center}
\end{table*}

\section{Related Work}

\label{related}

\paragraph{CS} Most prior work on CS focused on Language Identification (LID) \cite{SBM14,MAG16} and POS tagging \cite{SL08,VGS14,GGD16,BWF16}. In this work we focus on language modeling, which we find more challenging.

\paragraph{LM} Language models have been traditionally created by using the n-grams approach \cite{BDM92,CG96}. Recently, neural models gained more popularity, both using a feed-forward network for an n-gram language model \cite{BDVJ03,MB05} and using recurrent architectures that are fed with the sequence of words, one word at a time \cite{MKB10,ZSV14,GAG16,FGC16,MDB17}.

Some work has been done also on optimizing LMs for  ASR purposes, using discriminative training. Kuo et al. \shortcite{KF02}, Roark et al. \shortcite{RSC07} and Dikici et al. \shortcite{DS13} all improve LM for ASR by maximizing the probability of the correct candidates. All of them use candidates of ASR systems as ``negative" examples and train n-gram LMs or use linear models for classifying candidates. A closer approach to ours is used by Huang et al. \shortcite{HL18}. There, they optimize an RNNLM with a discriminative loss as part of training an ASR system. Unlike our proposed model, they still use the standard setting of LM. In addition, their training is coupled with an end-to-end ASR system, in particular, as in previous works, the ``negative" examples they use are candidates of that ASR system.

\paragraph{LM for CS}

Some work has been done also specifically on LM for code-switching. In Chan et al. \shortcite{CCCL09}, the authors compare different n-gram language models, Vu et al. \shortcite{VLW12} suggest to improve language modeling by generating artificial code-switched text. Li and Fung \shortcite{LF12} propose a language model that incorporates a syntactic constraint and combine both a code-switched LM and a monolingual LM in the decoding process of an ASR system. Later on, they also suggest to incorporate a different syntactic constraint and to learn the language model from bilingual data using it \cite{LF14}. Pratapa et al. \shortcite{PBC18} also use a syntactic constraint to improve LM by augmenting synthetically created CS sentences in which this constraint is not violated. Adel et al. \shortcite{AVK13} introduce an RNN based LM, where the output layer is factorized into languages, and POS tags are added to the input. In Adel et al. \shortcite{AVS13}, they further investigate an n-gram based factorized LM where each word in the input is concatenated with its POS tag and its language identifier. Adel et al. \shortcite{AKT14,AVK15} also investigate the influence of syntactic and semantic features in the framework of factorized language models. Sreeram and Sinha \shortcite{SS17} also use a factorize LM with the addition of POS tags. Baheti et al. \shortcite{CBS17} explore several different training protocols for CS LM and find that fine-tuning with CS data after pretraining on monolingual data works best.
Finally, another line of works suggests using a dual language model, where two monolingual LMs are combined by a probabilistic model \cite{GPJ17,GPJ18}.

No standard benchmark or evaluation setup exists for CS LM, and most previous works use an ASR system as part of their evaluation setup. This makes comparison of methods very challenging. Indeed, all the works listed above use different setups and don't compare to each other, even for works coming from the same group. We believe the evaluation setup we propose in this work and our English-Spanish dataset, which is easy to replicate and decoupled from an ASR system, is a needed step towards meaningful comparison of CS LM approaches.  

\section{Conclusions and Future Work}

We consider the topic of language modeling for code-switched data. We propose a novel ranking-based evaluation method for language models, motivated by speech recognition, and create an evaluation dataset for English-Spanish code-switching LM (SpanglishEval).	

We further present a discriminative training for this ranking task that is intended for ASR purposes. This training procedure is not bound to probability distributions, and uses both positive and negative training sentences. This significantly improves performance. Such discriminative training can also be applied to monolingual data.

Finally, we verify the effectiveness of the training protocol for CS LM presented in Baheti et al. \shortcite{CBS17}: pre-training on a mix of monolingual sentences, followed by fine-tuning on a code-switched dataset. This protocol significantly outperforms various baselines. Moreover, we show that the less code-switched training data we use, the more effective it is to incorporate the monolingual data. 

Our proposed evaluation framework and dataset will facilitate future work by providing the ability to meaningfully compare the performance of different methods to each other, an ability that was sorely missing in previous work.

\section*{Acknowledgments}

The work was supported by The Israeli Science Foundation (grant number 1555/15).

\bibliographystyle{acl_natbib}
\bibliography{bib_full}

\newpage

\appendix

\section{Creating Evaluation Dataset -- Implementation Details}

Our implementation is based on the Carmel FST toolkit.\footnote{\texttt{https://www.isi.edu/licensed-sw/} \texttt{carmel/}} We create an FST for converting a sentence into a sequence of phonemes, and its inverse FST. The words to phoneme mapping is based on pronunciation dictionaries, according to the language tag of each word in the sentence.

We use The CMU Pronouncing Dictionary\footnote{\texttt{http://www.speech.cs.cmu.edu/cgi-bin/} \texttt{cmudict}} for English and a dictionary from CMUSphinx\footnote{\texttt{https://sourceforge.net/projects/} \texttt{cmusphinx/files/Acoustic\%20and\%} \texttt{20Language\%20Models/Spanish/}} for Spanish. As the phoneme inventories in the two datasets do not match, we map the Spanish phonemes to the CMU dict inventory using a manually constructed mapping.\footnote{The full mapping from Spanish to English: ch-CH, rr-R, gn-NG, a-AA, b-B, b-V, e-EY, d-D, d-DH, g-G, f-F, i-IY, k-K, j-H, m-M, n-N, l-L, o-OW, p-P, s-S, r-R, u-UW, t-T, y-Y, x-S, x-SH, x-K S, x-H, z-TH, z-S, ll-L Y, ll-SH.  We thank Kyle Gorman for helping with the mapping.}

To favor frequent words over infrequent ones, we add unigram probabilities to the edges of the transducer (taken from googlebooks unigrams\footnote{\texttt{http://storage.googleapis.com/books/} \texttt{ngrams/books/datasetsv2.html}}). We filter some  words that produce noise (for example, single letter words that are too frequent). When creating a monolingual sentence, we use an FST with the words of that language only.

As many phoneme sequences in Spanish do not produce English alternatives (and vice versa) we allow minor changes in the phoneme sequences between the languages. Specifically, we create a small list of similar phonemes (such as "B" and "V"),\footnote{The full list of similar phonemes: OW - UW, 	AA - EY, L - M,	N - M, M - L, B - P, B - V, V - F, T - D, K - G, S - Z, S - TH, Z - TH, SH - ZH} and generate an FST that for each phoneme allows changing it to one of its alternatives or dropping it with low probability.

Since using the whole sentence has higher chances of encountering words that are not included in the dictionaries, we only convert a sampled part of the gold sentence when creating a code-switched alternative. This also results in alternatives with higher similarity to the gold sentence. However, when creating a monolingual alternative (i.e. a Spanish alternative to an English gold sentence), we have no choice but to use the whole sentence.

\section{Data}

\subsection{Code-switching corpus}

We pre-processed the Bangor Miami Corpus by lower-casing it and tokenizing using the spaCy tokenizer.\footnote{\texttt{https://spacy.io}}  We did not reduce the vocabulary size which was quite small to begin with (13,914 words). After preprocessing, we got 45,621 sentences with 322,044 tokens. 

\subsection{Monolingual corpora}

This data from the OpenSubtitles2018 corpus~\cite{T09} comes pre-tokenized. We pre-processed it by lower-casing, removing parenthesis and their contents, and removing hyphens from beginning of sentences.

We use 1M lines from each language, resulting in 7,501,714 tokens in English and 6,566,337 tokens in Spanish. We have 45,280 words in the English vocabulary and 50K words in the Spanish one (reduced from 83,615).

\section{Architecture and Training Details}

The LSTM has a hidden layer of dimension 650. The input embeddings are of dimension 300. We use auto-batching with batches of size 20. We optimize with SGD and learning rate of 10, reducing it by a factor of 2.5 at the end of each epoch with no improvement. We also use clipping gradients of 1, and weight decay of $10^{-5}$. We initialize the parameters of the LSTM to be in the range of $[-0.05,0.05]$. We also use word dropout with the rate of 0.2. We set the dropout in our LSTM \cite{GAG16} to 0.35. We train for 40 epochs and use the best model on the dev set.

\end{document}